\documentclass[10pt,twocolumn,letterpaper]{article}

\usepackage[pagenumbers]{cvpr} 

\usepackage{graphicx}
\usepackage{amsmath}
\usepackage{amssymb}
\usepackage{booktabs}

\usepackage[pagebackref,breaklinks,colorlinks]{hyperref}

\usepackage[capitalize]{cleveref}
\crefname{section}{Sec.}{Secs.}
\Crefname{section}{Section}{Sections}
\Crefname{table}{Table}{Tables}
\crefname{table}{Tab.}{Tabs.}

\def\cvprPaperID{0000} 
\def\confName{NULL}
\def\confYear{0000}

\begin{document}

\title{Consistent Arbitrary Style Transfer}

\author{Xuan Luo$^1$\\
\and
Zhen Han$^2$\\
Xi'an Jiaotong University$^1$\\
\and
lingkang Yang$^1$\\
Wuhan University$^2$\\
\and
Lingling Zhang$^1$\\
}
\maketitle

\begin{abstract}
Recently, attentional arbitrary style transfer methods have been proposed to achieve fine-grained results, which manipulates the point-wise similarity between content and style features for stylization. However, the attention mechanism based on feature points ignores the feature multi-manifold distribution, where each feature manifold corresponds to a semantic region in the image. Consequently, a uniform content semantic region is rendered by highly different patterns from various style semantic regions, producing inconsistent stylization results with visual artifacts. We proposed the progressive attentional manifold alignment (PAMA) to alleviate this problem, which repeatedly applies attention operations and space-aware interpolations. The attention operation rearranges style features dynamically according to the spatial distribution of content features. This makes the content and style manifolds correspond on the feature map. Then the space-aware interpolation adaptively interpolates between the corresponding content and style manifolds to increase their similarity. By gradually aligning the content manifolds to style manifolds, the proposed PAMA achieves state-of-the-art performance while avoiding the inconsistency of semantic regions. Codes are available at \url{https://github.com/computer-vision2022/PAMA}. 
\end{abstract}

\section{Introduction}
\label{sec:intro}

Neural style transfer aims at rendering a content image with style patterns from a reference image. The pioneering style transfer algorithm is proposed by Gatys \etal \cite{o1,o2}, which iteratively optimizes an image with perceptual losses. To accelerate the stylization process, a bunch of feed-forward network based approaches \cite{f1,f2,f3,f4,f5,f6,f7,f8,f9} flourished. These methods render the content image with a single forward propagation but cannot generalize well to unseen styles. Arbitrary style transfer methods \cite{a1, a2, a3, a5, a6} further extend the feed-forward network to arbitrary styles. These flexible yet efficient methods have received widespread attention from academics and industrial.

Recently, style attentional network (SANet)\cite{att1} and its following works \cite{att2, att3} have achieved state-of-the-art performance. With a learnable kernel, these methods compute pair-wise similarities to generate the attention map, which serves as a per-point feature transformation for fine-grained stylization. Still, the attentional style transfer methods are imperfect: they tend to stylize the semantic regions inconsistently. As demonstrated in the first row of \cref{fig:fig1}, the sky region is rendered by highly different style patterns from various semantic regions, causing visual artifacts. According to \cite{mani1}, each semantic region corresponds to a feature manifold, and the feature follows a multi-manifold distribution. To render a content region consistently, features from it should only be stylized by those from the most related style manifold. However, since the per-point transformation failed to capture the manifold distributions, features of the sky manifold are stylized independently, leading to chaotic results.

To address this problem, we want the attention module to learn a manifold-aware measurement, which consistently matches features between related content and style manifolds. Unfortunately, since the content semantic and style semantic are inherently dissimilar, their manifold distributions are heterogeneous. It is difficult for the attention module to learn the desired measurement. Therefore, we adopt the manifold alignment to reveal the cross manifold correspondence. We align each content manifold to its most related style manifold, thus increasing the structural similarity between them. Afterward, features from the corresponding content and style manifolds are close in space, making the attention module match them consistently. The existing manifold alignment style transfer method \cite{mani1} employs a global channel transformation for stylization. Instead of aligning the related content and style manifolds individually, this method aligns the multi-manifold distributions as a whole. As a result, it is not suitable to work with the attention module.

In this paper, we proposed the progressive attentional manifold alignment (PAMA) framework, which performs attention operations and space-aware interpolations multiple times. The proposed PAMA dynamically aligns each of the content manifolds to their most related style manifolds, enabling the consistent attention mechanism between regions. Firstly, the attention operation is employed to rearrange the style features according to the spatial structure of content features. By matching content and style features, the related content manifolds and style manifolds correspond on the feature map. Afterward, the space-aware interpolation fuses the corresponding manifolds with adaptive weights. The interpolation can increase the structural similarity of the corresponding manifolds dynamically, making the attention module easier to match features between them. However, a single alignment cannot build a strong enough correspondence between manifolds for the attention module. Therefore we repeat the manifold alignment process multiple times and employ multistage loss functions. An image reconstruction loss is also used to maintain the shared space for manifold alignment. Our contributions can be summarized as:

\begin{itemize}
\item[$\bullet$]We proposed a new arbitrary style transfer framework named PAMA, which gradually aligns content manifolds to style manifolds with the attention mechanism for consistent stylization between semantic regions. 
\item[$\bullet$]A multistage loss function is designed to enable progressive manifold alignment while preserving the regional consistency. We also adopt an image reconstruction loss to maintain the shared space for manifold alignment.
\item[$\bullet$]Experiments show that the proposed framework can generate fine-grained stylization result in real-time (101 fps for 512px images on Tesla V100 GPU).  
\end{itemize}

\section{Related Works}
\label{sec:rel}
\subsection{Arbitrary Style Transfer}
The goal of arbitrary style transfer is to generate stylization results in real-time with arbitrary content-style pairs. The mainstream arbitrary style transfer algorithms can be divided into two groups: the global transformation based and local patch based. The global transformation based method utilizes the global statistic for feature transformation. One of the representative methods is AdaIN\cite{a1}, which forces the mean and variance of content features to be the same as the style features for stylization. The WCT\cite{a2} changes the covariance matrix of content features with whitening and coloring transformation. For Linear\cite{a4}, it learns the feature transformation matrix explicitly with a subnet.  Although effective, the global transformation is vulnerable to local content distortions. 

For the local patch based methods, they manipulate the feature patches for stylization. The style swap proposed by Chen \etal \cite{patch1} is the earliest patch based method, which swaps the content patches with the most similar style patches. The DFR\cite{patch2} and AAMS\cite{patch3} further extend this method with the global statistics. These local patch based methods can preserve the content structure fairly well but tend to trigger visual artifacts. Recently, the SANet\cite{att1} is proposed to match content and patches with a learnable similarity kernel dynamically. Then MANet\cite{att2} disentangles the content and style representation for attention mechanism. AdaAttN\cite{att3} enhances the attention mechanism with multi-layer features to reduce the content distortion. But as the attentional methods render the feature points independently, they fail to capture the multi-manifold distribution of features, producing inconsistent stylization results.
\subsection{Manifold Alignment}
Manifold alignment aims at revealing the relationship of two datasets from different manifold distributions. These algorithms learn to project the two datasets to a shared subspace, where the correspondence of two datasets is established, and their manifold structures are preserved. Existing manifold alignment methods can be divided into semi-supervised methods\cite{m1,m2,m3} and unsupervised methods\cite{m4,m5,m6}. Huo \etal introduce manifold alignment methods to the style transfer community\cite{mani1}, which makes an assumption that the image features follow a multi-manifold distribution. This method aligns the content manifolds to style manifolds with a global channel transformation. However, It cannot increase the similarity of related manifolds in a targeted manner, and thus not suitable for consistent attention mechanism between semantic regions.


\section{Method}
\subsection{Pre-Analysis}
In this section, we will explain the inconsistent stylization phenomenon from a manifold perspective. According to \cite{mani1}, the image features follow a multi-manifold distribution that each semantic region corresponds to a manifold. If there were $m$ semantic regions in the content image, we can divide the content features $F_c$ into $m$ subsets corresponding to the manifolds:
\begin{equation}
\label{eq:eq1}
F_c = \cup_{i=1}^m F_{c,i} \ , F_c \subseteq \mathbb{R}^{C \times H_cW_c}, F_{c,i} \subseteq \mathbb{R}^{C\times M_i}
\end{equation}
where $F_{c,i}$ denotes the $i$-th subset (corresponding to the $i$-th manifold) which consists of $M_i$ features. The style features $F_s$ follows the same definition: 
\begin{equation}
\label{eq:eq2}
F_s = \cup_{i=1}^n F_{s,i} \ , F_s \subseteq \mathbb{R}^{C \times H_sW_s}, F_{s,i} \subseteq \mathbb{R}^{C\times N_i}
\end{equation}
here we have defined the content features and style features in a manifold perspective.

The inconsistency happens because the per-point rendering of the attention mechanism neglects the multi-manifold distribution. Suppose the $p$-th content manifold is most related to the $q$-th style manifold, which means that features from $F_{c,p}$ should only match with features from $F_{s,q}$ in attention. For any content feature $f_c^x$ from $F_c$, if it belongs to $F_{c,p}$, we can write the attention operation as:  
\begin{equation}
\label{eq:eq3}
\begin{aligned}
\hat f_s^x & = \sum_{i=1}^{H_sW_s} sim(f_c^x, f_s^i)f_s^i \\
& = \sum_{i=1}^{N_q} sim(f_c^x, f_{s,q}^i)f_{s,q}^i+\eta(f_c^x, F_s)
\end{aligned}
\end{equation}
where $f_s^i$ denotes the $i$-th feature from $F_s$, and the $f_{s,q}^i$ refers to the $i$-th feature from the $q$-th subset of $F_s$. The learned similarity kernel of attention mechanism abbreviates as $sim(\cdot, \cdot)$. The $\eta(\cdot, \cdot)$ is the mismatch term measuring the inconsistency. This equation suggests that features from a content manifold should only be stylized by features from the most related style manifold, and any out-of-manifold rendering is considered as mismatch. If the mismatch term $\eta$ is relatively high, the stylization result tend to be inconsistent. To reduce the mismatch term and enable consistent attention module, we proposed the progressive attentional manifold alignment (PAMA) to align the content manifolds to their most related style manifolds.


\subsection{Overall Framework}

\begin{figure}[t]
  \centering
   \includegraphics[width=\linewidth]{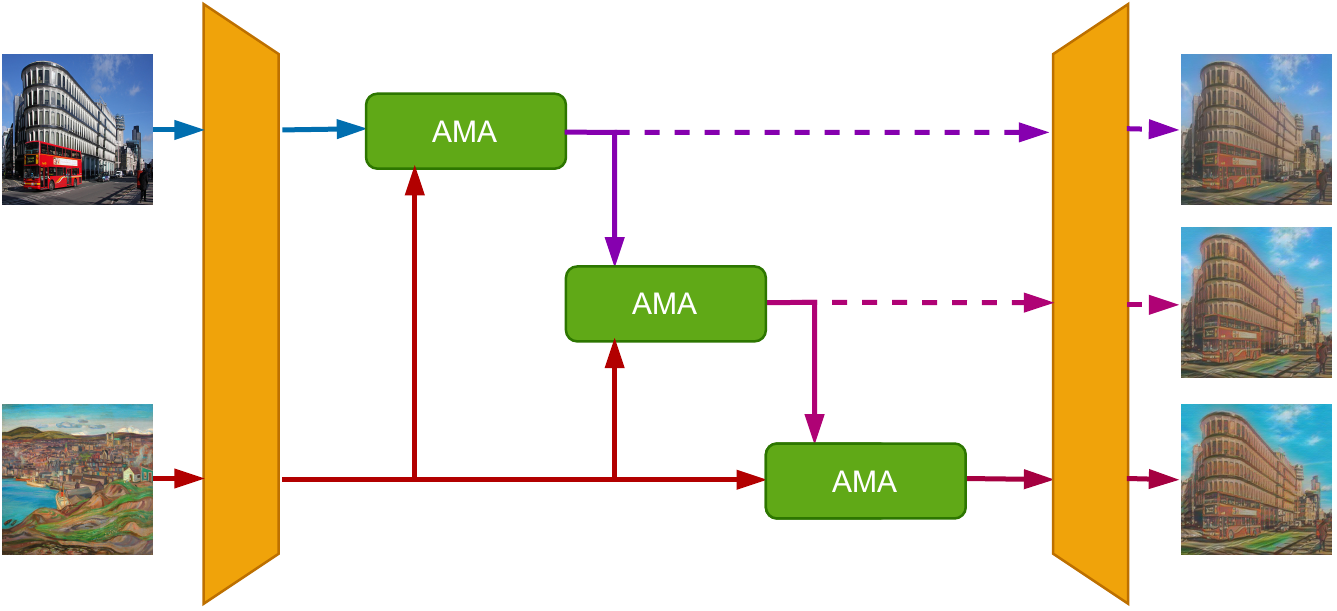}

   \caption{The architecture of our network. The content manifolds is gradually aligned to the style manifolds with three independent attentional manifold alignment (AMA) blocks. The dash lines are only forwarded during training to generate the intermediate results for loss calculation. }
   \label{fig:fig2}
\end{figure}

\cref{fig:fig2} shows the proposed progressive attentional manifold alignment (PAMA) framework. Our method uses a pre-trained VGG \cite{vgg} network to encode the content image $I_c$ and style image $I_s$ to obtain the $ReLU4\_1$ features $F_c$ and $F_s$. The attentional manifold alignment (AMA) consists of an attention module and a space-aware interpolation module. Passed through three AMA blocks, the content feature $F_c$ gradually fuses style information, during which the content manifolds are aligned to the style manifolds. In this way, the attention mechanism can capture the manifold distributions and render the semantic regions consistently. Finally, the aligned content feature will be fed into the decoder to generate the stylized image. The structure of the decoder is symmetric to the encoder. Notice that the intermediate stylized features are only decoded during training to calculate the multistage perceptual loss (dash lines in \cref{fig:fig2}).


\subsection{Loss Function}

\begin{figure}[t]
  \centering
   \includegraphics[width=0.8\linewidth]{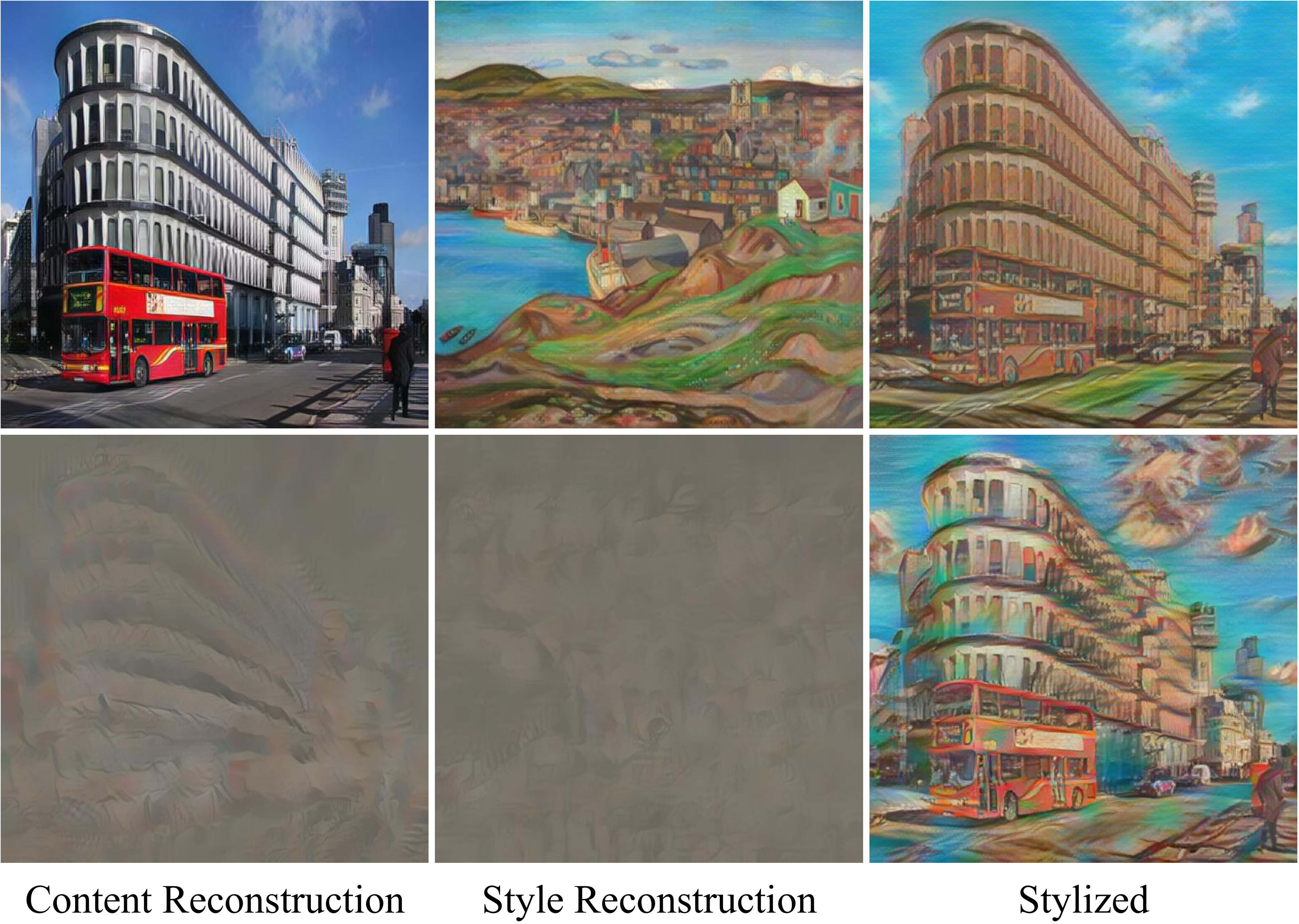}

   \caption{Attention mechanism will change the original VGG\cite{vgg} space. The first row is generated by the proposed PAMA, the second row shows the result of SANet\cite{att1}.}
   \label{fig:fig3}
\end{figure}

The attentional transformation in existing methods like \cite{att1, att2, att3} will change the original VGG\cite{vgg} space of style feature. \cref{fig:fig3} demonstrates that the decoder of SANet failed to parse the original content and style feature in the VGG space. This characteristic is devastating for our progressive alignment that the attention module needs to learn a cross space similarity measurement, making the inconsistency problem even worse. To constraint all of the features in the VGG space, we employed an image reconstruction loss: 
\begin{equation}
\label{eq:eq4}
\begin{aligned}
& L_{rec} = \lambda(\|(I_{rc}-I_c)\|_2 + \|(I_{rs}-I_s)\|_2) + \\
& \sum_{i}(\|\phi_i(I_{rc})-\phi_i(I_c)\|_2 + \|\phi_i(I_{rs})-\phi_i(I_s)\|_2)
 \end{aligned}
\end{equation}
where $I_{rc}$ and $I_{rs}$ are the content and style image reconstructed from VGG features, and the $\lambda$ is a constant weight. The $\phi_i(I)$ refers to the $ReLU\_i\_1$ layer VGG feature of image $I$. Since this loss forces the decoder to reconstruct VGG features, all features between the encoder and decoder are restricted in the VGG space. In the manifold alignment perspective, this loss function maintains the shared space for the alignment between content and style manifolds.

For the progressive manifold alignment, the overall loss function consists of multiple stages. In each stage, the loss is a weighted summation:
\begin{equation}
\label{eq:eq5}
L = \sum_{i=1}(\lambda_{ss}^i L_{ss} + \lambda_r^i L_r + \lambda_m^i L_m + \lambda_{h}^i L_{h})+L_{rec}
\end{equation}
where the $\lambda_x^i$ denotes the weight for $L_x$ in the $i$-th stage. The $L_{ss}$ is the content loss, while the $L_r$, $L_m$, and $L_h$ serve as the style losses. The initial value of the content weight $\lambda_{ss}^1$ is relatively high to preserve the content manifold structure. In the next stages, the content weight decreases gradually to encourage more vivid style patterns.

Our content loss is based on the structure self-similarity descriptor\cite{remd1} between the content feature $F_c$ and the VGG feature of stylized image $F_{cs}$:
\begin{equation}
 \label{eq:eq6}
L_{ss} = \frac{1}{H_c W_c} \sum_{i,j} |\frac{D_{ij}^c}{\sum_i D_{ij}^c}-\frac{D_{ij}^{cs}}{\sum_j D_{ij}^{cs}}|
\end{equation}
where $D_{ij}^c$ and $D_{ij}^{cs}$ are the pairwise cosine distance matrices of $F_c$ and $F_{cs}$ respectively. This loss function is effective for manifold structure preservation, because it regulates the feature correlation within manifolds to be invariant. 

Following the setting of \cite{remd1, remd2} we adapts the relaxed earth mover distance (REMD):
\begin{equation}
\label{eq:eq7}
L_{r} = \max (\frac{1}{H_sW_s}\sum_i \min_j C_{ij}, \frac{1}{H_cW_c}\sum_j \min_i C_{ij})
\end{equation}
where the $C_{ij}$ denotes the pair-wise cosine distance matrix between $F_{cs}$ and $F_s$. As suggested in \cite{remd2}, this loss function optimizes along the manifold surface of style features, which works well with our manifold alignment process. We also added the moment matching loss to regularize the magnitude of features:
\begin{equation}
\label{eq:eq8}
L_m = \|\mu_{cs} - \mu_s\|_1 + \|\Sigma_{cs} - \Sigma_s\|_1
\end{equation}
where the $\mu$ and $\Sigma$ denotes the mean and covariance matrix of feature vectors.

Although our proposed PAMA generates high-quality stylized images, it sometimes outputs color-mixed images. The self-similarity loss causes this limitation, for it forces the attention mechanism rendering a region overly uniform, making it mix the style patterns. To fix it, we referred to the differentiable color histogram loss proposed in HistoGAN\cite{hist}:
\begin{equation}
\label{eq:eq9}
L_{h} = \frac{1}{\sqrt 2} \| H_s^{1/2} - H_{cs}^{1/2} \|_2
\end{equation}
where the $H$ refers the color histogram feature, the $H^{1/2}$ denotes the element-wise square root.  At the expense of a little consistency, this loss function can reduce the color mixing problem.


\begin{figure}[t]
  \centering
   \includegraphics[width=\linewidth]{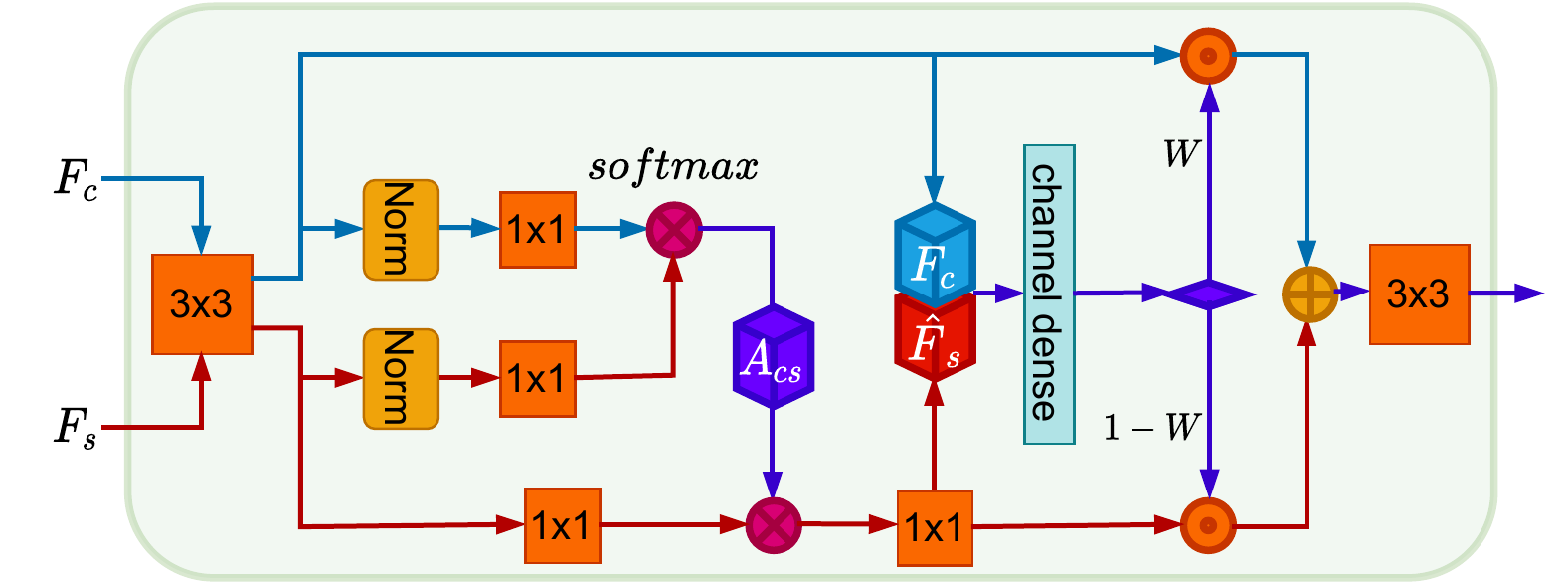}

   \caption{The attentional manifold alignment (AMA) block. The attention module generates the attention map $A_{cs}$ to rearrange the style feature. The space-aware interpolation dense the channel information to obtain the adaptive weight $W$, which is applied to interpolate between the content feature $F_c$ and the rearranged style feature $\hat F_s$.}
   \label{fig:fig4}
\end{figure}

\subsection{Attentional Manifold Alignment Block}
\cref{fig:fig4} demonstrates the attention manifold alignment block, which consists of an attention module and a space-aware interpolation module. 

\textbf{Attention Module}. In the attention module, the content and style features are firstly normalized and embedded to compute the attention map:
\begin{equation}
\label{eq:eq10}
A_{cs} = softmax(f(Norm(F_c))^T \otimes g(Norm(F_s)))
\end{equation}
where the $f(\cdot)$ and $g(\cdot)$ denote 1x1 convolution blocks for feature embedding, the $Norm(\cdot)$ refers to the mean-variance normalization, and the $\otimes$ is the matrix multiplication. The attention map contains the pair-wise similarities between the content features and style features. Then the attention map serves as an affine transformation to spatially rearrange the style features: 
\begin{equation}
\label{eq:eq11}
\hat F_s = \theta(h(F_s)^T \otimes A_{cs})
\end{equation}
again the $g(\cdot)$ and $\theta(\cdot)$ are the 1x1 convolutions for feature embedding.

\begin{figure}[t]
  \centering
   \includegraphics[width=\linewidth]{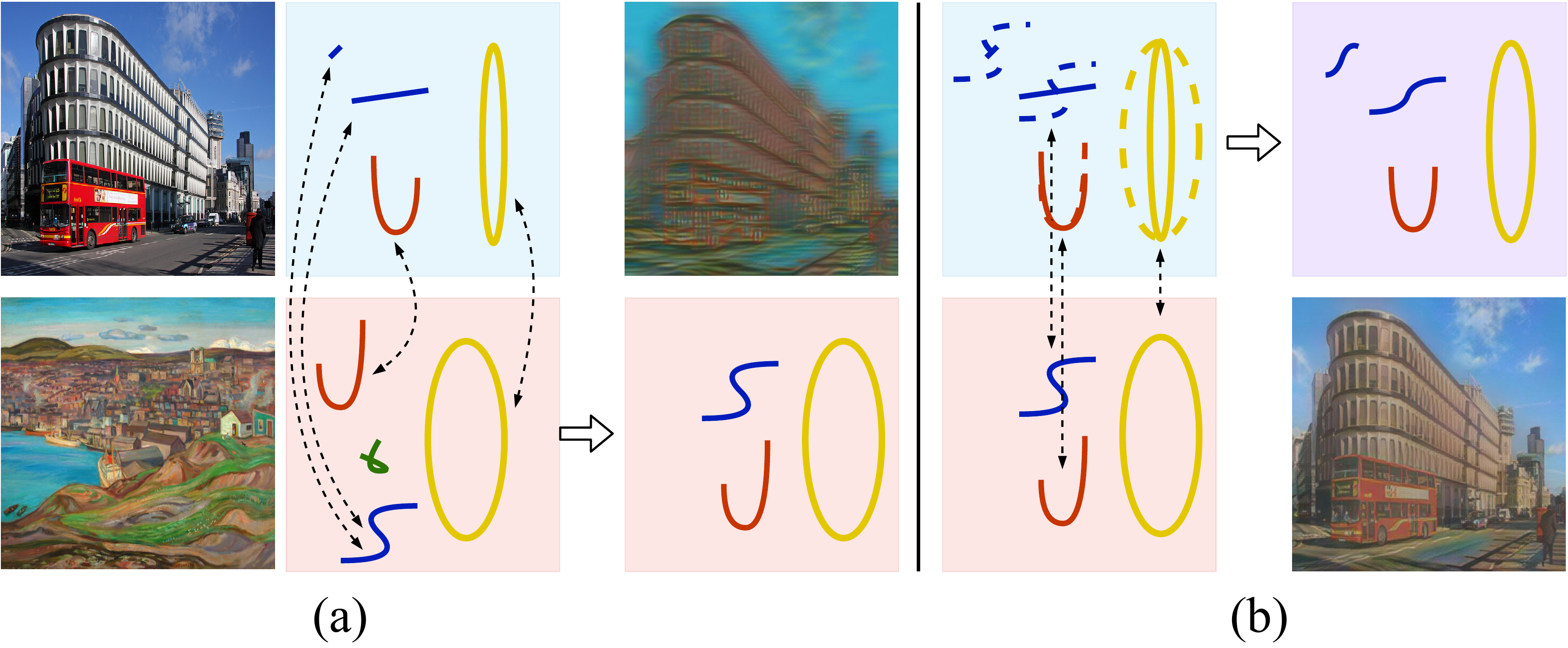}

   \caption{The manifold alignment process in the first stage. (a) The attention operation to find the correspondence between manifolds; (b) The interpolation to increase the structural similarity between corresponding manifolds.}
   \label{fig:fig5}
\end{figure}

The attention module is exactly the same as the one in \cite{att1, att2}, which produces inconsistent results. In the attention module of PAMA, we aim at finding the correspondence between related content and style semantic regions while minimizing the mismatch term $\eta$ in \cref{eq:eq3}. Therefore, we adopt a high initial value of self-similarity loss (\cref{eq:eq6}) and decrease it gradually during the multi-stage manifold alignment. In the first stage, the high self-similarity loss encourages the manifold structure of the rearranged style features $\hat F_s$ to be the same as the content feature $F_c$, thus forcing the attention module to match the features consistently. \cref{fig:fig5} (a) shows that the style features $F_s$ are rearranged according to the spatial structure of content features $F_c$, and the related semantic regions correspond on the feature map. With this correspondence, the space-aware interpolation module can increase the structural similarity between the related manifolds. In the next stages, even if we decrease the self-similarity loss for more vivid style patterns, the attention module will reveal the relationship between manifolds and rendering the features consistently. We have verified this in the experiment part with \cref{fig:fig8}.

\textbf{Space-aware Interpolation Module.} This module adaptively interpolates between $F_c$ and $\hat F_s$ with regional information. Initially, the channel dense operation applies convolution kernels of different scale on the concatenated feature to summarize multi-scale regional information:
\begin{equation}
 \label{eq:eq12}
W = \frac{1}{n}\sum_{i=1}^n \psi_i ([F_c, \hat F_s]) 
\end{equation}
where $\psi_i (\cdot)$ represent the $i$-th convolution kernel, and the $[\cdot, \cdot]$ denotes the channel concatenation operation. The concatenated feature can help us to identify the differences between the corresponding content and style manifolds, figuring out the local inconsistencies triggered by the attention module. This learnable channel dense operation outputs the scalar spatial weights $W \in \mathbb{R}^{H \times W}$, which is used for interpolation:
\begin{equation}
 \label{eq:eq13}
F_{cs} = W \odot F_c +(1-W) \odot \hat F_s
\end{equation}
where the $\odot$ refers to the dot production. Different from the ordinary feature fusion in attentional methods like \cite{att1, att2, att3}, the space-aware interpolation fuses features in the same space. Therefore the interpolated content feature will not suffer from degradation (see \cref{fig:fig3}). As you can see in \cref{fig:fig5} (b), the interpolation fully solved the local distortion. Moreover, the affinity between the corresponding manifolds are increased, making it easier for the attention module to render semantic regions consistently. Then the stylized feature $F_{cs}$ will be feed into the next attentional alignment block for further refinement.

\section{Experiment}
\subsection{Implementation Details}
Our proposed PAMA is trained with content images from COCO\cite{coco} and style images from wikiart\cite{wiki}. Features from $ReLU\_3\_1$, $ReLU\_4\_1$, and $ReLU\_5\_1$ are used to compute the self-similarity loss $L_{ss}$, the REMD loss $L_r$, and the moment matching loss $L_m$. For self-similarity loss, we set its weights $\lambda_{ss}^1$, $\lambda_{ss}^2$, $\lambda_{ss}^3$ to 12, 9, 7 respectively. All weights of the style loss terms $L_r$ and $L_m$ are set to 2. These terms are normalized with the summation of their weights (\eg $\frac{12}{12+2+2}$ for $\lambda_{ss}^1$). The weights of color histogram loss $\lambda_h^1$,  $\lambda_h^2$, $\lambda_h^3$ are set to 0.25, 0.5, and 1. The $\lambda$ for the reconstruction loss in \cref{eq:eq5} is 50.  We use the Adam\cite{adam} as our optimizer with learning rate of 0.0001 and momentum of 0.9. During training, we take 8 content-style image pairs as a batch. The smaller dimension of content and style images are rescaled to 512, and then we randomly crop a 256x256 patch for efficient training. In the testing phase, our fully convolutional network can tackle images with any size.

\begin{figure*}[t]
  \centering
   \includegraphics[width=\linewidth]{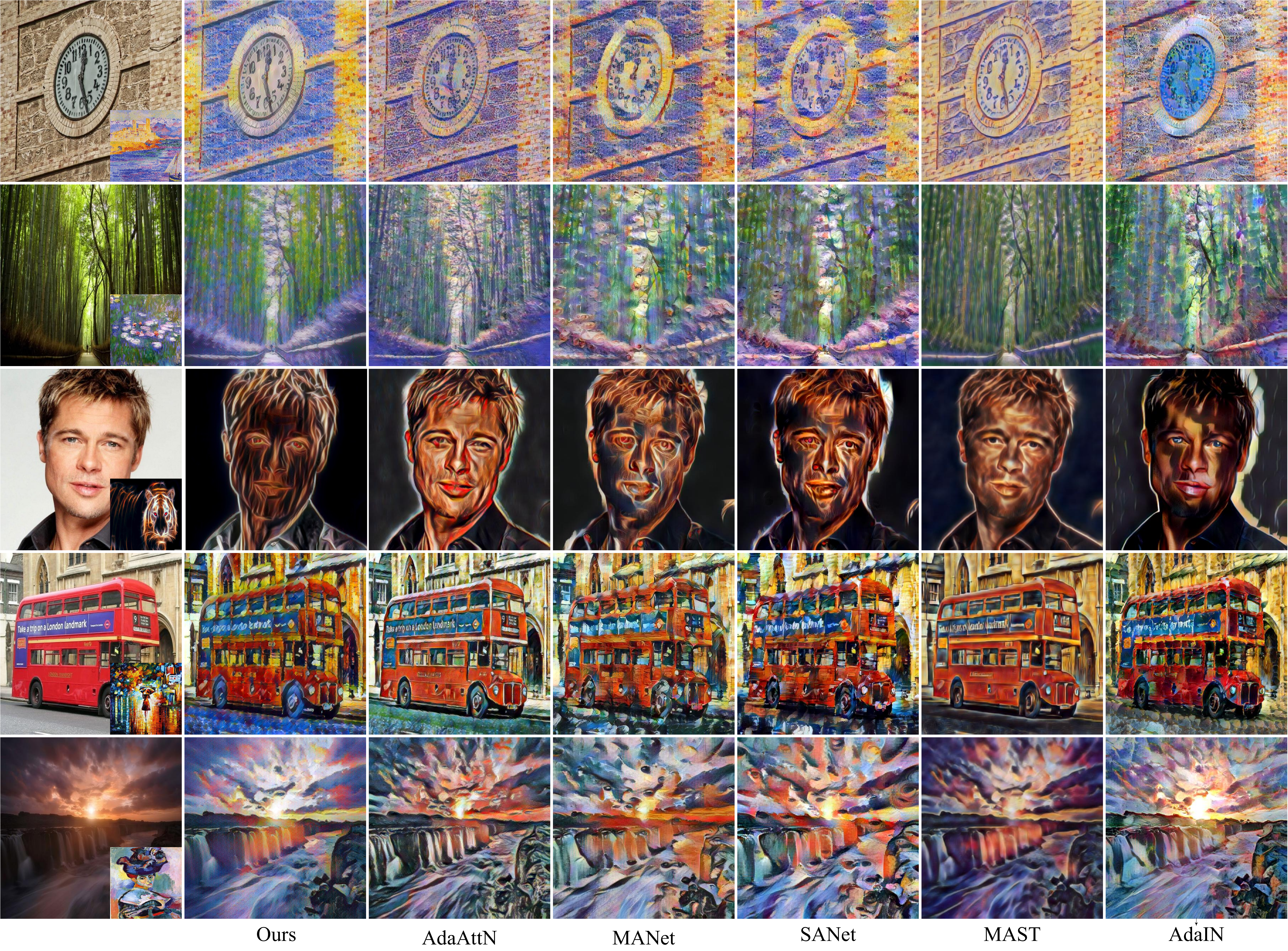}
   \caption{Qualitative Comparison.}
   \label{fig:fig6}
\end{figure*}

\subsection{Comparison with Prior Arts}
To evaluate the propose approach, we compared it with other arbitrary style transfer methods, including the pioneering method AdaIN\cite{a1}, the first manifold-alignment-based method MAST \cite{mani1}, and the attention-based methods \cite{att1, att2, att3}. 

\textbf{Qualitative Comparison}
In \cref{fig:fig6} we show the comparison results. AdaIN\cite{a1} adjusts the mean and variance of the content feature globally for stylization. For its global transformation neglects the local details, AdaIN tends to bring distortions (1st, 2nd, 3rd, 5th rows) and visual artifacts(2nd, 3rd rows). MAST\cite{mani1} is an arbitrary style transfer algorithm based on manifold alignment, which can stylize the semantic regions consistently. Rather than treating different manifolds dynamically, MAST aligns the multi-manifold distribution as a whole with a global channel transformation. This makes the MAST suffers from style degradation (2nd, 4th, 5th rows) and blurry (2nd, 5th rows). SANet\cite{att1} and MANet\cite{att2} use the attention mechanism to transform the deep features point-wise. They produce fine-grained results with vivid style patterns but are still not free from the distortion problem (2nd, 4th rows) and the visual artifacts (3rd, 5th rows). Because these attentional methods render the feature points independently without considering the manifold distribution, their semantic regions are not consistent (1st, 2nd, 3rd rows). The AdaAttN\cite{att3} extends the attention mechanism with shallow features and achieves better content preservation. However, as you can see in the 3rd and 4th rows, increasing content protection capabilities sacrifices style patterns: the color distribution and brushstrokes of AdaAttN are quite different from the style image.  The AdaAttN is also affected by inconsistency (1st, 2nd rows) and visual artifacts (5th row). In the 1st and 2nd row, because the PAMA has aligned the content and style manifolds for the attention module, it can produce much more consistent results. In the 3rd and 5th rows, you can see that our methods also generate significantly fewer visual artifacts. The proposed PAMA achieves a better balance between content and style while increasing the consistency in semantic regions.

\begin{figure}[t]
  \centering
   \includegraphics[width=\linewidth]{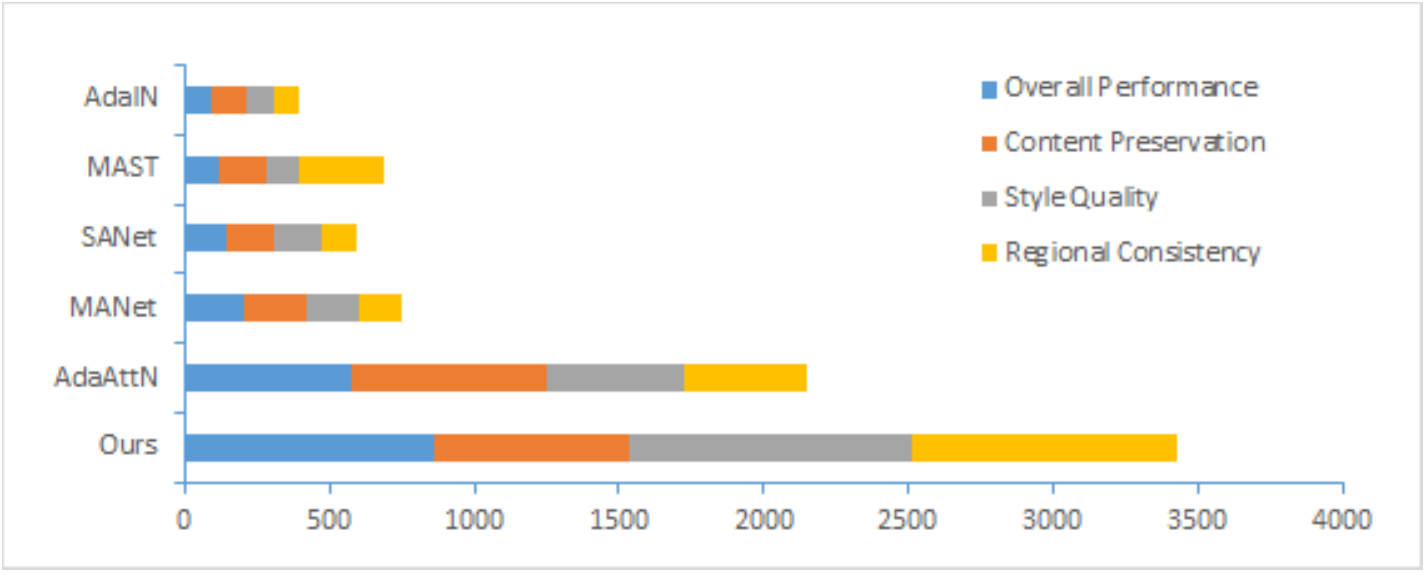}
   \caption{User preference result of AdaIN\cite{a1}, MAST\cite{mani1}, SANet\cite{att1}, MANet\cite{att2}, and AdaAttN\cite{att3}. }
   \label{fig:fig7}
\end{figure}

\textbf{User Study.}
We use 25 content images and 25 style images to generate 625 stylization results in total for each method. Then 20 samples are randomly drawn from the 625 results and distributed to users. For each sample, users are asked to choose their favorite stylization results among all methods. We used 4 evaluation indicators: content preservation, style quality, regional consistency, and overall performance. We collect votes from 100 users and plot \cref{fig:fig7}. The result demonstrates that our method produces results with significantly higher consistency and overall performance. 

\textbf{Efficiency.}
In \cref{tab:tab1}, we compare the running time of our method with other baselines at 256px and 512px. All of the methods are evaluated on a server with an Intel Xeon Gold 5118 CPU @ 2.30GHz and an NVIDIA V100 PCIe 32G GPU. The results are the average running time of 100 image pairs. Thanks to the lightweight design of the attentional manifold alignment block, even if we align the manifold multiple times, our method can still generate artworks in real-time (101 fps at 512px).

\begin{table}
  \centering
  \begin{tabular}{p{2cm} p{2cm} p{2cm}}
    \toprule
    Method & Time (256px) & Time (512px) \\
    \midrule
    AdaIN & 2.813ms  &  2.912ms \\
    MAST & 1488ms & 2743ms \\
    SANet & 4.792ms  &  6.351ms \\
    MANet & 8.202ms  &  8.992ms \\
    AdaAttN & 19.76ms  &  22.52ms\\
    Ours & 8.527ms &  9.871ms\\
    \bottomrule
  \end{tabular}
  \caption{Running time comparison.}
  \label{tab:tab1}
\end{table}


\subsection{Verify the Effectiveness}

\textbf{Attention Operation.}
\begin{figure}[t]
  \centering
   \includegraphics[width=\linewidth]{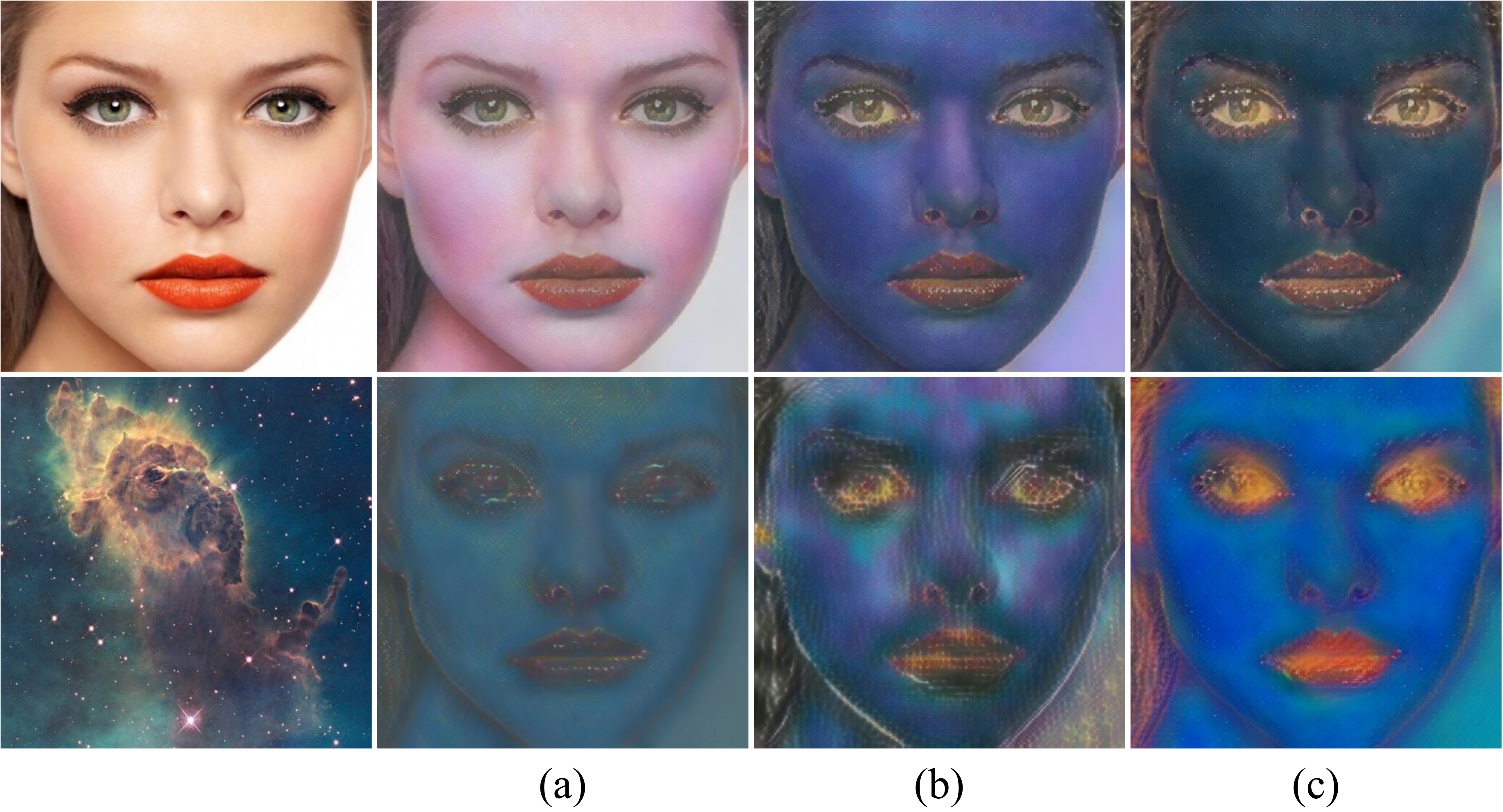}
   \caption{Multi-stage attention results. Images in the first row are stylization results. Images from the second row are the decoded results of rearrange style feature $\hat F_s$. (a) Results from the first stage; (b) Results from the second stage; (c) Results from the third stage.}
   \label{fig:fig8}
\end{figure}
To analyze the behavior of attention operation during the progressive manifold alignment process, we visualize the rearranged style feature $\hat F_s$ in different stages. For the first stage, a relatively high self-similar loss is used for training, which regulates the manifold structure of the content image to be invariant. As we can see in \cref{fig:fig8} (a) that the rearranged style feature is well aligned with the content structure. Although the result is consistent, the attention module fails to capture the manifold distributions, rendering all of the regions uniformly. Here, the consistency is provided by the self-similarity loss rather than the attention module. In the second stage, for the content manifold are fused with related style information in the first stage, the attention module can distinguish different content manifolds and stylize them with features from various style manifolds. In the final stage, as shown in \cref{fig:fig8}, the content manifolds are well aligned with the most similar style manifolds, and the content semantic regions are rendered consistently by the style manifolds. The above proves the effectiveness of the manifold alignment method for the consistent attention mechanism.

\begin{figure}[t]
  \centering
   \includegraphics[width=\linewidth]{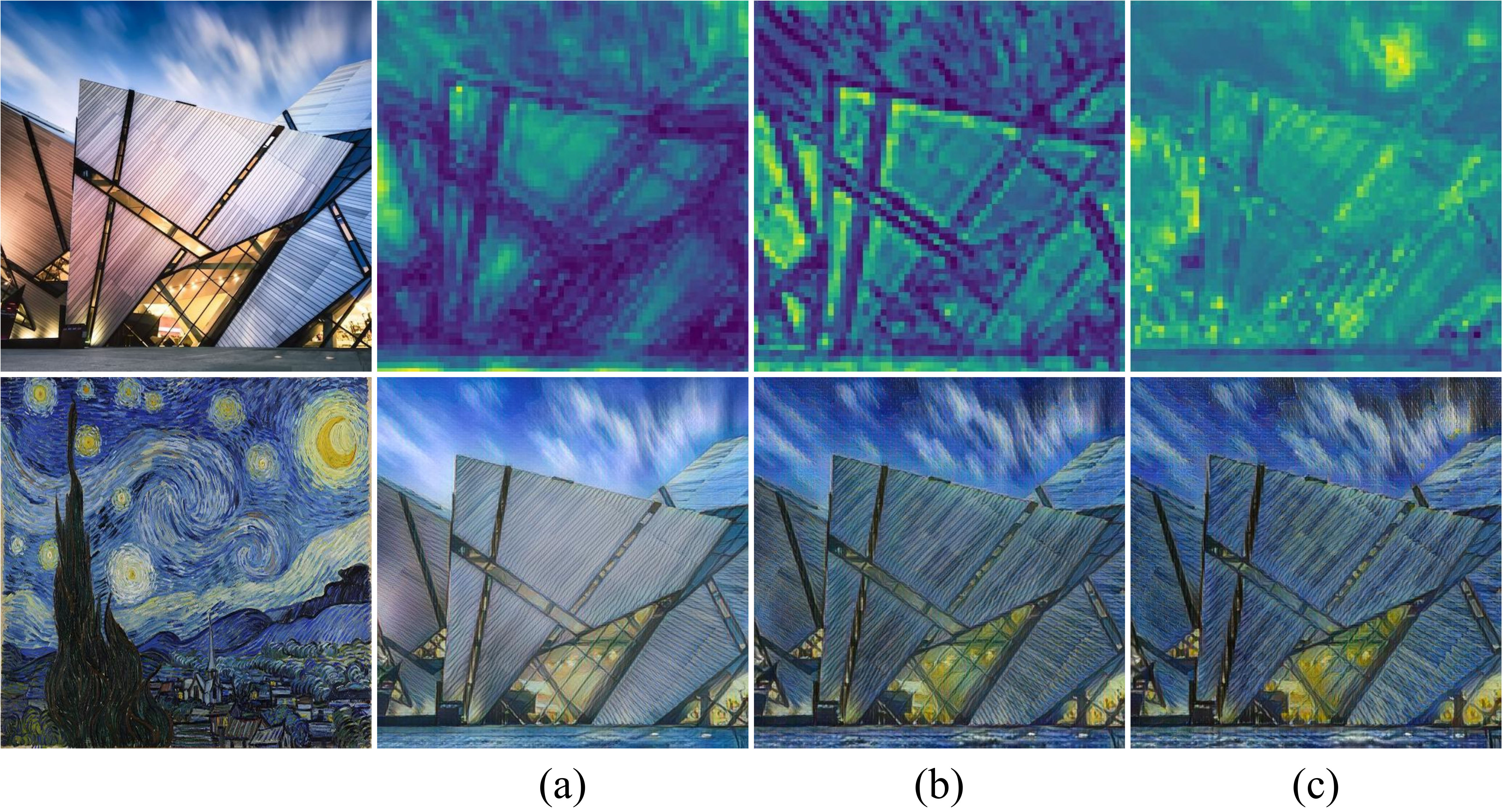}
   \caption{Space-aware Interpolation results. The first row is the visualization of the adaptive weight $W$, where the yellow pixels denote higher stylization, and the blue pixels denote higher content preservation. The second row is the interpolation results of the first row respectively. (a) Interpolation of the first stage; (b) Interpolation of the second stage; (c) Interpolation of the third stage. }
   \label{fig:fig9}
\end{figure}
\begin{figure*}[t]
  \centering
   \includegraphics[width=\linewidth]{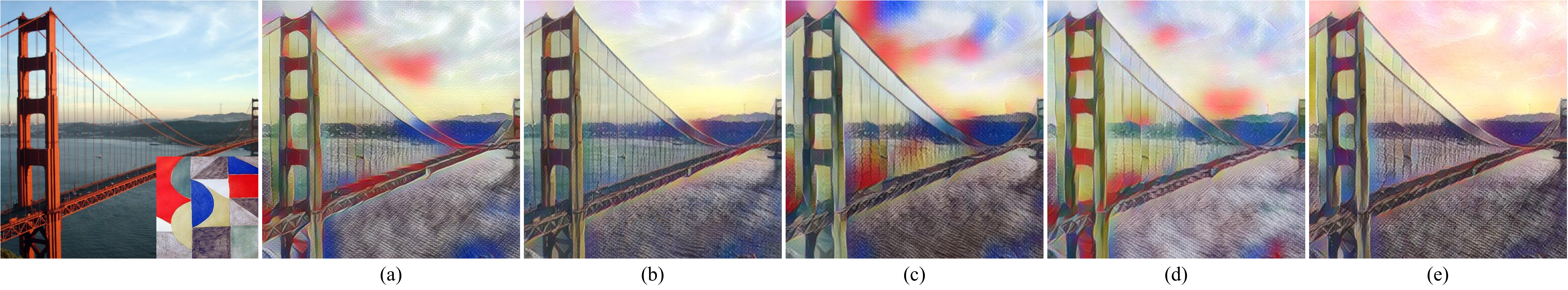}
   \caption{The effectiveness of the self-similarity loss and color histogram loss. (a) Original PAMA; (b) High self-similarity loss; (c) Low self-similarity loss; (d) High histogram loss; (e) low histogram loss.}
   \label{fig:fig10}
\end{figure*}
\begin{figure}[t]
  \centering
   \includegraphics[width=\linewidth]{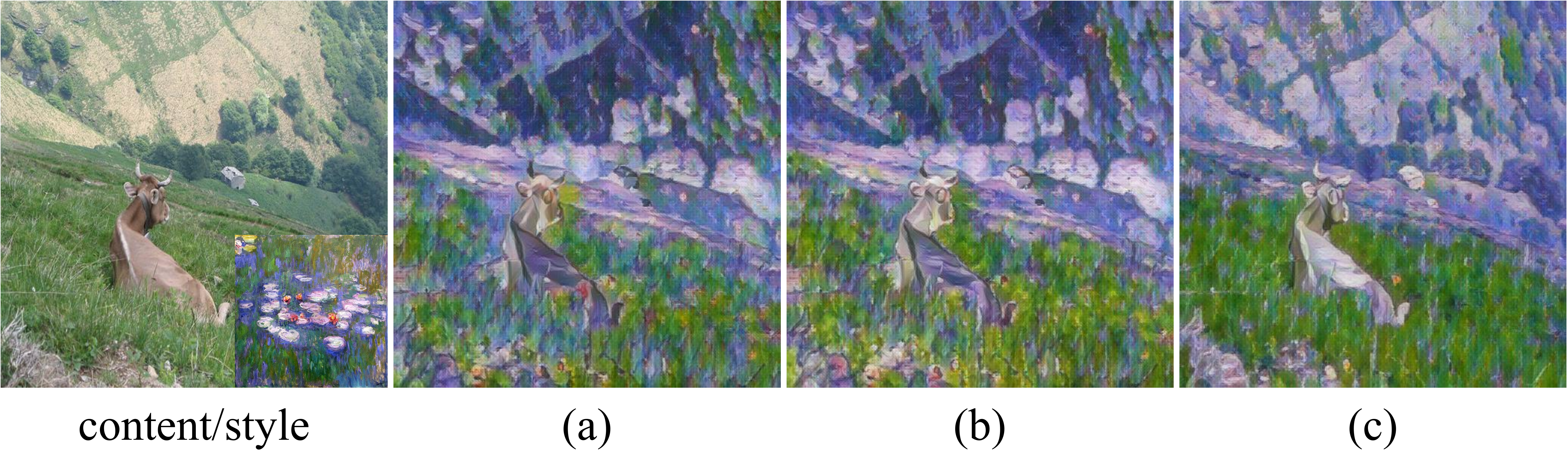}
   \caption{The effectiveness of multi-stage manifold alignment. (a) Single-stage manifold alignment; (b) Two-stage manifold alignment; (c) Three-stage manifold alignment. }
   \label{fig:fig11}
\end{figure}

\textbf{Space-aware Interpolation.}
This part verifies the effectiveness of the space-aware interpolation of all three manifold alignment stages. \cref{fig:fig9} demonstrates that the space-aware interpolations are sensitive to edge information and tend to preserve content structures around the edges. This can help the network to remove the local inconsistency (or distortions) in salient areas. Also, the interpolation module has learned to detect the uniform regions and render them with higher strength. In this way, we can add style patterns without hurting the manifold structure significantly. For the reason that we decrease the self-similarity content loss gradually during manifold alignment, the interpolation module fuses more style information in the latter stages (more yellow pixels in \cref{fig:fig9}), producing results with vivid style patterns.

\textbf{Loss Analysis.}
In Figure 10 we explore the effect of different loss weights, including the self-similarity loss $\lambda_{ss}$ and the color histogram loss $\lambda_h$. \cref{fig:fig10} (a) is the original result of the proposed PAMA. The sky part has been rendered with red and yellow patterns chaotically. This is due to the color histogram loss forcing the color distribution of the result to be the same as the style image. \cref{fig:fig10} (b) shows the result with high self-similarity loss, its weights are $\lambda_{ss}^{1}=20$, $\lambda_{ss}^2=15$, $\lambda_{ss}^3=10$. After zooming in, you can find that the sky part is rendered uniformly, and the slings of the suspension bridge are well preserved. The self-similarity loss protects the structure of content regions. For \cref{fig:fig10} (c), its $\lambda_{ss}^1=10$, $\lambda_{ss}^2=6$, $\lambda_{ss}^3=4$. Although contains more vivid colors, the high style loss has introduced distortions and inconsistencies. In terms of the color histogram loss, \cref{fig:fig10} (d) uses $L_h^1=0.5$, $L_h^2=1$, $L_h^3=2$. The high color histogram loss leads to more appropriate color rendering but also brings inconsistent color blocks in the image. $\cref{fig:fig10}$ (e) is the result with color histogram weights of $L_h^1=0.125$, $L_h^2=0.25$, $L_h^3=0.5$. This example has the highest consistency in that there is no fluctuation in color in most of the parts. However, the overly uniform leads to the color mixing problem, the color of the sky is mixed into a pinkish color. In conclusion, the balance of self-similarity loss and color histogram loss controls the balance of consistency and color distribution. 

\textbf{Multistage Manifold Alignment.}
To investigate the behavior of multistage manifold alignment, we trained another two networks with single-stage manifold alignment and two-stage manifold alignment, respectively. For the single-stage manifold alignment network, we trains it with content loss weight $L_{ss}^1=7$, REMD loss weight $L_r^1=2$, moment loss weight $L_m^1=2$, and histogram loss weight $L_h^1=1$. It is exactly the weight of the last stage of PAMA. \cref{fig:fig11} (a) demonstrates that the grassland is stylized by green and purple patterns inconsistently. Single-stage manifold alignment is not enough to reveal the manifold structure. The two-stage manifold alignment network is trained with the same loss function and weights in the last two stages of PAMA. The consistency of the grassland improves, but local inconsistency still exists. \cref{fig:fig11} (c) is the result generated by the proposed PAMA, and you can see a monotonous increase of consistency.

\section{Conclusion}
In this paper, we proposed the progressive attention manifold alignment framework (PAMA) to alleviate the inconsistency problem and improve the quality of stylization. PAMA uses the attention mechanism to reveal the correspondence between the most related content and style manifolds and then applies the space-aware interpolation to fuse the related manifolds adaptively. By performing this process multiple times, the structural similarity between the most related content and style manifolds increases, making it easier for the attention mechanism to match features consistently between them. The attention mechanism can capture the multi-manifold distribution of features to generate high-quality consistent results.
\clearpage
{\small
\bibliographystyle{ieee_fullname}
\bibliography{egbib}
}

\end{document}